\pdfoutput=1
\documentclass[11pt]{article}

\usepackage{acl}
\usepackage{times}
\usepackage{latexsym}
\usepackage[T1]{fontenc}
\usepackage[utf8]{inputenc}
\usepackage{inconsolata}
\usepackage{graphicx}
\usepackage[dvipsnames]{xcolor}

\usepackage{comment}
\usepackage{paralist}
\usepackage{booktabs}
\usepackage{tcolorbox}
\usepackage{amsmath}
\usepackage{shellesc}

\usepackage{microtype}

\title{Does Self-Consistency Improve the Recall of Encyclopedic Knowledge?}

\author{Sho Hoshino, Ukyo Honda, Peinan Zhang \\
CyberAgent \\
\texttt{\{hoshino\_sho,honda\_ukyo,zhang\_peinan\}@cyberagent.co.jp}
}

\begin{document}
\maketitle
\begin{abstract}
While self-consistency is known to improve performance on symbolic reasoning, its effect on the recall of encyclopedic knowledge is unclear due to a lack of targeted evaluation grounds.
To address this, we establish such a knowledge recall split for the popular MMLU benchmark by applying a data-driven heuristic from prior work.
We validate this split by showing that the performance patterns on the symbolic reasoning and knowledge recall subsets mirror those of GSM8K and MedMCQA, respectively.
Using this solid ground, we find that self-consistency consistently improves performance across both symbolic reasoning and knowledge recall, even though its underlying CoT prompting is primarily effective for symbolic reasoning.
As a result, we achieve an 89\% accuracy on MMLU, the best performance to date with the use of GPT-4o.
\end{abstract}

\section{Introduction}
The chain-of-thought prompting \citep[CoT;][]{wei2022chain,nye2022show} has become the de facto standard for achieving the best performance on large language model (LLM) benchmarks \cite{openai2024gpt4technicalreport,geminiteam2024gemini}.
Nevertheless, it is widely held that CoT is primarily effective for tasks requiring symbolic reasoning.
\citet{sprague2025to} reported that on the Massive Multitask Language Understanding (MMLU) benchmark \cite{hendrycks2021measuring}, 95\% of the performance gain from CoT is attributed to questions involving symbolic reasoning.

\begin{figure}[t]
\centering
\includegraphics[width=\linewidth]{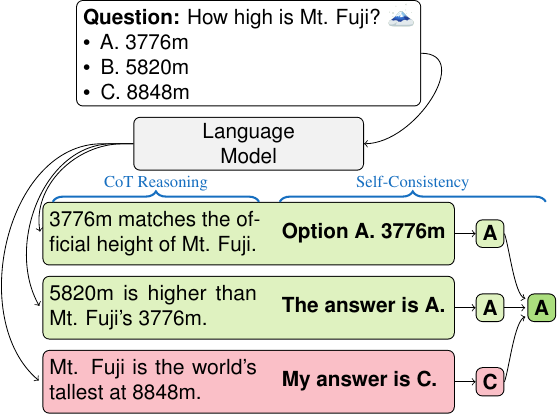}
\caption{Illustrating how self-consistency can mitigate incorrect reasoning path for a knowledge recall question.}
\label{fig:eyecatch}
\end{figure}

Building on top of CoT, self-consistency \citep[SC;][]{wang2023selfconsistency} further improves performance by sampling multiple CoT reasoning paths and selecting the most consistent answer.
However, as self-consistency is fundamentally an extension of CoT, it is unclear whether self-consistency also improves performance on non-math questions that involve the recall of encyclopedic knowledge (hereafter, \emph{knowledge recall}\footnote{We followed the term coined by \citet{chung2024scaling}.}), such as \textit{``How high is Mt. Fuji?''} shown in Figure~\ref{fig:eyecatch}.
To investigate this gap, we address two primary research questions:
Does knowledge recall benefit from multiple reasoning paths (\textbf{RQ1})?
If so, how does self-consistency improve knowledge recall with multiple samples (\textbf{RQ2})?

Our assessment reveals that self-consistency improves both the symbolic reasoning and knowledge recall performance over vanilla CoT, despite its premise.
To enable this analysis, we apply a subject-level split to the MMLU dataset based on a prior heuristic.
We validate this split by showing that the performance patterns on the symbolic reasoning and knowledge recall subsets mirror those of prototypical benchmarks, GSM8K \cite{{cobbe2021trainingverifierssolvemath}} and MedMCQA \cite{pmlr-v174-pal22a}, respectively.
As a result, we achieve an 89\% accuracy on MMLU, the best performance to date with the use of GPT-4o.

We further analyzed the mechanism behind this unexpected improvement.
Our quantitative analysis found that the agreement among answers serves as a reliable confidence score, providing evidence that self-consistency filters out incorrect paths by leveraging this signal.
Our qualitative analysis further illustrates how self-consistency stabilizes unreliable CoT reasoning by filtering out these incorrect paths, making it effective for both symbolic reasoning and knowledge recall.

\section{Data Splitting Methodology}
Since the MMLU benchmark \cite{hendrycks2021measuring} contains a wide range of 57 subjects, several categorization methods have been proposed.
Indeed, \citet{hendrycks2021measuring} defined their own subject-level categorization with ``supercategory'', such as STEM and humanities.
Their grouping was not designed to distinguish knowledge recall from symbolic reasoning.
For example, the subject of econometrics, which requires symbolic reasoning, is categorized as humanities.

A more targeted approach by \citet{sprague2025to} used the ''$=$'' sign as an instance-level cue for symbolic reasoning.
Their post-hoc analysis was model-dependent, as it relied on the cue's presence in either the question or the LLM's output.
While this analysis was post-hoc, the heuristic originated from a classifier trained to differentiate subjects, providing a basis for subject-level split.

Building on these insights, we apply \citeposs{sprague2025to} heuristic across all 57 MMLU subjects to create a stable, \textit{a priori} categorization independent of model outputs, as shown in Figure~\ref{fig:subject}.
We aggregated subjects guided solely by the presence of ``$=$'' in the questions and then propagated this classification within a discipline (e.g., from college math to elementary math).
Our application disentangles knowledge recall from symbolic reasoning in a ratio of about 2:1.

To validate our split, we select two prototypical benchmarks, including GSM8K, a math dataset, and MedMCQA, a medical dataset where symbolic reasoning cues are almost entirely absent (16 out of 4,183 instances).
We expect the performance patterns on our MMLU subsets to mirror those on these prototypical benchmarks.
A data-driven analysis in Appendix~\ref{sec:data-driven}, which shows a high overlap (AUC of 0.96) between our split and one based on CoT performance gains, provides further validation.

\begin{figure}[t]
\centering
\footnotesize
\begin{tcolorbox}[boxrule=0.5pt, top=0pt, bottom=0pt, left=0pt, right=0pt, colback=OliveGreen!35, title=Symbolic Reasoning Subjects]
Abstract Algebra, Business Ethics, College Chemistry, College Computer Science,
College Mathematics, College Physics, Conceptual Physics, Econometrics, Elementary Mathematics, Formal Logic, High School Chemistry, High School Computer Science, High School Macroeconomics, High School Mathematics, High School Microeconomics, High School Physics, Machine Learning, Professional Accounting
\end{tcolorbox}
\begin{tcolorbox}[boxrule=0.5pt, top=0pt, bottom=0pt, left=0pt, right=0pt, colback=Tan!15, title=Knowledge Recall Subjects]
Anatomy, Astronomy, Clinical Knowledge, College Biology, College Medicine, Computer Security, Electrical Engineering, Global Facts, High School Biology, High School European History, High School Geography, High School Government and Politics, High School Psychology, High School Statistics, High School US History, High School World History, Human Aging, Human Sexuality, International Law, Jurisprudence, Logical Fallacies, Management, Marketing, Medical Genetics, Miscellaneous, Moral Disputes, Moral Scenarios, Nutrition, Philosophy, Prehistory, Professional Law, Professional Medicine, Professional Psychology, Public Relations, Security Studies, Sociology, US Foreign Policy, Virology, World Religions
\end{tcolorbox}
\caption{Listing symbolic reasoning and knowledge recall subjects defined for the MMLU dataset.}
\label{fig:subject}
\end{figure}

\section{Experiments}
We first validate our proposed MMLU split to ensure it effectively separates symbolic reasoning and knowledge recall.
With the split validated, we then present our main results on the effectiveness of self-consistency, configured as follows.\footnote
{We provide full implementation details, including hyperparameters and data splitting, in Appendix~\ref{appendix:detail}.}

\subsection{Setup}

\paragraph{Data.}
We use the MMLU \cite{hendrycks2021measuring} and MedMCQA \cite{pmlr-v174-pal22a} for multiple-choice question answering \citep[MCQA;][]{balepur-etal-2025-best}, and GSM8K \cite{cobbe2021trainingverifierssolvemath} as an open-ended question answering prototype.

\paragraph{Metrics.}
To ensure a rigorous assessment, we evaluate using classification accuracy for the MCQA datasets, and a similar accuracy metric for GSM8K.
We have performed experiments using the zero-shot setting, as opposed to the few-shot setting \cite{chung2024scaling}.\footnote
{We also performed experiments with a similar few-shot setting, as discussed in Appendix~\ref{sec:vs-fewshot}.}

\begin{table*}[t]
\centering
\footnotesize
\begin{tabular}{@{}c@{}llllll@{}}
\toprule
& & \multicolumn{5}{c}{Accuracy (\%) $\uparrow$} \\
& & \multicolumn{3}{c}{MMLU test} & GSM8K test & MedMCQA valid \\
\cmidrule(l){3-5}
Prompt & \multicolumn{1}{c}{Sampling} & \multicolumn{1}{c}{All} & \multicolumn{1}{c}{Reasoning} & \multicolumn{1}{c}{Knowledge} \\
\midrule
DA & \multicolumn{1}{c}{nucleus} & 83.26 & 75.45 & 85.56 & 46.93 & 75.07 \\
CoT & \multicolumn{1}{c}{nucleus} & 87.86\phantom{$^*$}\scriptsize{(+4.60)} & 90.38\phantom{$^*$}\colorbox{OliveGreen!35}{\scriptsize{(+14.93)}} & 87.12\phantom{$^*$}\colorbox{Tan!15}{\scriptsize{(+1.56)}} & 84.23~\colorbox{OliveGreen!35}{\scriptsize{(+37.30)}} & 76.76~\colorbox{Tan!15}{\scriptsize{(+1.69)}} \\
\midrule
CoT & +\textsc{SC}~\scriptsize{($n$=5)} & \textbf{88.64}$^*$\scriptsize{(+5.38)} & \textbf{91.32}$^*$\colorbox{OliveGreen!35}{\scriptsize{(+15.87)}} & \textbf{87.85}$^*$\colorbox{Tan!15}{\scriptsize{(+2.29)}} & \textbf{84.31}~\colorbox{OliveGreen!35}{\scriptsize{(+37.38)}} & \textbf{77.67}~\colorbox{Tan!15}{\scriptsize{(+2.60)}} \\
CoT & +\textsc{SC}~\scriptsize{($n$=20)} & \textbf{88.93}$^*$\scriptsize{(+5.67)} & \textbf{91.94}$^*$\colorbox{OliveGreen!35}{\scriptsize{(+16.49)}} & \textbf{88.04}$^*$\colorbox{Tan!15}{\scriptsize{(+2.48)}} & \textbf{84.46}~\colorbox{OliveGreen!35}{\scriptsize{(+37.53)}} & \textbf{77.41}~\colorbox{Tan!15}{\scriptsize{(+2.34)}} \\
\bottomrule
\end{tabular}
\caption{
Performance of GPT-4o on MMLU, GSM8K, and MedMCQA.
The top section validates our MMLU split, which consists of the full test set (``All''), the symbolic reasoning subset (``Reasoning''), and the knowledge recall subset (``Knowledge'').
The bottom section shows that self-consistency (SC) consistently improves performance over the vanilla CoT baseline for both symbolic reasoning and knowledge recall, with these improvements highlighted in bold.
The asterisk ($^*$) denotes statistical significance ($p < 0.05$, detailed in Appendix~\ref{appendix:detail}).
}\label{tab:main}
\end{table*}

\paragraph{LLM and Prompt.}
We use GPT-4o version 2024-08-06, GPT-4o-mini version 2024-07-18 \cite{openai2024gpt4technicalreport}, and Qwen2.5-32B-Instruct \cite{qwen2025qwen25technicalreport}, without changing hyperparameters unless explicitly mentioned.\footnote{
Our main analysis focuses on GPT-4o for clarity.
Full results for all models are in Appendix~\ref{sec:full}.
}
We use two types of prompts, including the zero-shot CoT \cite{kojima2022large} and direct answer without CoT (abbreviated as DA).

\paragraph{Sampling.}
We use nucleus sampling \cite{holtzman2020curious} with top-$p$=0.9.
For the vanilla CoT baseline, the first sample is selected deterministically.
For the self-consistency, we varied the number of samples ($n$) from 3 to 20.

\subsection{Validation of the MMLU Split}
To validate our proposed split, we examine the performance difference between CoT and direct answer.
We expect symbolic reasoning to benefit significantly from CoT, while knowledge recall should show little to no gain.
The top section of Table~\ref{tab:main} confirms this.
The MMLU symbolic reasoning subset shows a substantial +14.93 point gain from CoT (75.45 to 90.38), while the knowledge recall subset shows only a modest +1.56 point gain (85.56 to 87.12).

This pattern is mirrored on our prototypical benchmarks, GSM8K (symbolic reasoning) and MedMCQA (knowledge recall).
GSM8K sees a massive +37.3 point gain from CoT (46.93 to 84.23), whereas MedMCQA sees a small +1.69 point gain (75.07 to 76.76).
The strong alignment between the MMLU splits and their corresponding prototypical benchmarks validates our split.

\subsection{Self-Consistency on Knowledge Recall}
Having validated our split, we now investigate whether knowledge recall benefits from multiple reasoning paths (\textbf{RQ1}).
The bottom section of Table~\ref{tab:main} shows that self-consistency consistently improves performance over vanilla CoT for both symbolic reasoning and, crucially, knowledge recall.
As a result, we achieved an overall accuracy of 89\% on the MMLU, the highest score to date using GPT-4o.\footnote
{Similar scores are already reported for different LLMs.}

This trend holds on our prototypical benchmarks as well.
As shown in Table~\ref{tab:main}, self-consistency provides gains for both GSM8K (symbolic reasoning) and MedMCQA (knowledge recall).
These results further validate the robustness of our findings.

\begin{figure}[t]
\centering
\includegraphics[width=\linewidth]{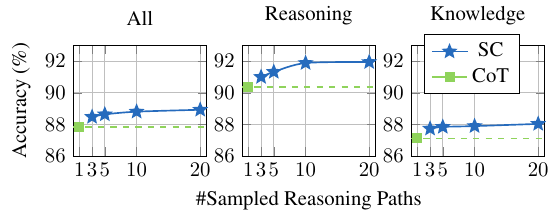}
\caption{
Comparing SC and vanilla CoT on MMLU using different numbers of samples from GPT-4o.
}
\label{fig:chart}
\end{figure}

As shown in Figure~\ref{fig:chart}, self-consistency outperformed vanilla CoT with no degradation observed, even when using different numbers of samples.

\section{Analysis}
To further investigate how self-consistency improves knowledge recall with multiple samples (\textbf{RQ2}), we present quantitative evidence for its core mechanism and then provide a qualitative illustration of the underlying reasoning patterns.

\subsection{Quantitative Evidence of the Mechanism}
\label{sec:quantitative}
We quantitatively test self-consistency's core mechanism, which relies on agreement among answers as a strong signal for correctness.
To do so, we formalize a confidence score as $s=\frac{\text{the count of the majority answer}}{\text{the number of valid answers}}$.
For example, from the three answers $\{\text{A}, \text{A}, \text{C}\}$ in
Figure~\ref{fig:eyecatch}, our confidence in the final answer A is calculated as
$s=\frac{2}{3}$.
This principle aligns with findings from a separate line of work showing that confidence metrics based on answer consistency are more reliable than than methods based on prediction logits \cite{yang-etal-2025-maqa}, especially since the first token may not indicate the final answer \cite{wang-etal-2024-answer-c}.
To validate that this score is a meaningful signal, we measure its Pearson correlation coefficient \cite[$\rho$;][]{pearson1895rho} with the correctness of the predictions.

\begin{table}[t]
\centering
\footnotesize
\begin{tabular}{@{}c@{}lccc@{}}
\toprule
& & \multicolumn{3}{c}{Pearson's $\rho$ [-1, 1] $\uparrow$} \\
& & \multicolumn{3}{c}{MMLU test} \\
\cmidrule(l){3-5}
Prompt & \multicolumn{1}{c}{Sampling} & All & Reasoning & Knowledge \\
\midrule
\textsc{CoT} & +\textsc{SC}~\scriptsize{($n$=5)} & 0.40 & 0.43 & 0.40 \\
\textsc{CoT} & +\textsc{SC}~\scriptsize{($n$=20)} & 0.42 & 0.46 & 0.42 \\
\bottomrule
\end{tabular}
\caption{
Correlation between prediction correctness and confidence based on answer agreement.
}\label{tab:confidence}
\end{table}

As summarized in Table~\ref{tab:confidence}, using the count of the majority answer as a confidence score showed a moderate correlation with the correctness of the LLMs' predictions.
This finding provides quantitative evidence for self-consistency's core mechanism, confirming that majority voting effectively filters out incorrect reasoning paths by leveraging this signal.

\subsection{Qualitative Illustration}
To complement the quantitative finding, we analyzed typical CoT reasoning paths from the MMLU development set to understand why different samples produce different answers, as shown in Figure~\ref{fig:example}.
Unlike symbolic reasoning that involves multi-hop reasoning, one would expect knowledge recall to be a single deduction that does not require multiple reasoning paths.

Unexpectedly, we found that multiple reasoning paths exist because the model generates different plausible justifications for the same question, even for knowledge recall.
For example, both reasoning paths in Figure~\ref{fig:example} contain seemingly plausible justifications (e.g., \textit{``A situation analysis would typically be conducted here''}), despite leading to different answers.
Our observations indicate that self-consistency does not alter how LLMs work, but rather stabilizes unreliable CoT reasoning by filtering out these incorrect paths.
By selecting the paths that lead to the most consistent final answer, the technique makes CoT effective for both symbolic reasoning and knowledge recall.

\begin{figure}[t]
\centering
\footnotesize
\begin{tcolorbox}[boxrule=0.5pt, top=0pt, bottom=0pt, left=0pt, right=0pt, colback=white]
Question: At which stage in the planning process would a situation analysis be carried out? \\
A. Defining the program \\
B. Planning the program \\
C. Taking action and implementing ideas
\end{tcolorbox}
\begin{tcolorbox}[boxrule=0.5pt, top=0pt, bottom=0pt, left=0pt, right=0pt, colback=LimeGreen!25, title=Correct Reasoning Path]
A. Defining the program:
This stage involves identifying the overall goals and objectives of the program. It sets the foundation for what the program aims to achieve.
\textbf{Situation analysis is crucial here} because it helps understand the current environment, needs, and gaps, which are essential to defining the goals and objectives accurately. \\
Therefore, the correct answer is: A. Defining the program
\end{tcolorbox}
\begin{tcolorbox}[boxrule=0.5pt, top=0pt, bottom=0pt, left=0pt, right=0pt, colback=OrangeRed!25, title=Incorrect Reasoning Path, ]
2. **Planning the program**: In this stage, specific plans are formulated based on the initial definition of the program. This is where detailed analysis and assessments are often conducted to inform the planning process. \textbf{A situation analysis would typically be conducted here} to understand the current state, identify opportunities and threats, and develop strategies based on this analysis. \\
Thus, the correct option is: B. Planning the program
\end{tcolorbox}
\caption{An example of typical reasoning paths for a multiple-choice question involving knowledge recall. In this example, the option A is the correct answer.}
\label{fig:example}
\end{figure}

\subsection{Cost-Performance Trade-off}
While self-consistency consistently improves knowledge recall, this gain comes with a practical trade-off because the inference cost scales linearly with the sample size $n$.
As illustrated in Figure~\ref{fig:chart}, increasing the number of samples yields diminishing returns in performance.
To quantify this trade-off, achieving a +0.73 point gain on the MMLU knowledge recall subset ($n=5$) requires approximately five times the computational cost compared to vanilla CoT ($n=1$).

\section{Related Work}
\citet{wang2023selfconsistency} proposed self-consistency and performed experiments on arithmetic and commonsense reasoning benchmarks.
After that, \citet{chung2024scaling} conducted additional experiments including the MMLU benchmark, which is perhaps the most related work.
However, their study did not provide a detailed breakdown of MMLU results by subject, leaving the performance on knowledge recall unclear.

\citet{gema-etal-2025-done} reported that more than 9\% of MMLU examples are considered incorrect due to errors in the dataset creation.
To mitigate such errors, several fixes to the MMLU dataset have been proposed, including MMLU-Redux \cite{gema-etal-2025-done}, MMLU-Pro \cite{wang2024mmlupro}, and MMLU-CF \cite{zhao-etal-2025-mmlu}.
Nevertheless, the subject-level split we use is orthogonal to these instance-level fixes.

\section{Conclusion}
To investigate self-consistency's performance on the recall of encyclopedic knowledge, we applied a subject-level split to MMLU based on a prior heuristic.
Our assessment on MMLU demonstrated that self-consistency yields consistent improvements over vanilla CoT for both symbolic reasoning and knowledge recall.
As a result, we achieved 89\% accuracy, the highest score to date on the MMLU with the use of GPT-4o.

\section*{Limitations}

\paragraph{On the Granularity of the Data Split}
One might critique our subject-level split between symbolic reasoning and knowledge recall as a coarse-grained approximation.
We agree that any such partition is an imperfect proxy, and there is no perfect approach, as even instance-level classifications can be ambiguous.
Our subject-level split is no exception, as some subjects contain a mix of question types.
However, our goal was to move beyond treating MMLU as a monolithic benchmark.
Therefore, we provide a stable, \textit{a priori} split as a solid ground, offering a clear improvement over the status quo.

\paragraph{On the Practical Costs of Self-Consistency}
One could argue that the performance gain from self-consistency comes with additional costs.
We admit these practical drawbacks, as self-consistency requires multiple samples of CoT reasoning paths, which means longer runtime and higher cost compared to vanilla CoT.
In this study, however, our focus was to establish a targeted evaluation ground.
By creating a principled \textit{a priori} split, we established the solid ground needed to answer the fundamental question of whether self-consistency improves the recall of encyclopedic knowledge.

\paragraph{On the Scope of Task Formats}
While our experiments on MMLU involve multiple-choice question answering to ensure a rigorous assessment via classification accuracy, our scope is not limited to this task format.
In fact, we included GSM8K as a open-ended question answering prototype.
Furthermore, future work can address arbitrary tasks with universal self-consistency \cite{chen2024universal}.


\bibliography{custom}

\clearpage
\appendix

\section{Setup Details}
\label{appendix:detail}

\paragraph{Data Splitting.}
For the MMLU dataset, we use dev split (285 instances) for development and test split (14,042 instances) for testing.
For the GSM8K dataset, we use test split (1,319 instances) for testing.
For the MedMCQA dataset, we use validation split (4,183 instances) for testing, instead of test split, because the latter's gold labels are not publicly available.

\paragraph{Hyperparameters.}
We set max tokens to 20 for DA and 1,000 for CoT, respectively, because the GPT-4o models tend to output longer, surpassing the 128 tokens previously used for the GPT-3 models \cite{wang2023selfconsistency}.

\paragraph{Prompts.}
\label{appendix:prompt}
Figure~\ref{fig:prompt} lists our prompt used as direct answer and chain-of-thought.
In our preliminary study, we observed significant performance gains with the combinations of general and negative instructions proposed in Appendix~\ref{sec:implementation}.

\paragraph{Statistical Testing.}
We assessed statistical significance using paired bootstrap resampling \cite{koehn-2004-statistical}.
The improvements from self-consistency are statistically significant ($p < 0.05$) on the MMLU benchmark across the evaluated models, with the exception of the knowledge recall subset for Qwen2.5-32B-Instruct at $n=5$.
The performance gains on GSM8K and MedMCQA were consistent in direction, while they did not reach statistical significance in all configurations, likely due to the smaller size of these test sets.

\paragraph{Licenses.}
We performed experiments using the MMLU dataset \cite{hendrycks2021measuring} released under the MIT license\footnote{ \url{https://huggingface.co/datasets/cais/mmlu}}, the GSM8K dataset \cite{{cobbe2021trainingverifierssolvemath}} released under the MIT license\footnote{\url{https://huggingface.co/datasets/openai/gsm8k}}, the MedMCQA dataset \cite{pmlr-v174-pal22a} released under the Apache License 2.0\footnote{\url{https://huggingface.co/datasets/openlifescienceai/medmcqa}}, GPT-4o API \cite{openai2024gpt4technicalreport} released under a proprietary license via Azure OpenAI Service\footnote{\url{https://azure.microsoft.com/en-us/products/ai-services/openai-service}}, and Qwen2.5-32B-Instruct \cite{qwen2025qwen25technicalreport} released under the Apache License 2.0\footnote{\url{https://huggingface.co/Qwen/Qwen2.5-32B-Instruct}}.

\begin{figure}[t]
\centering
\footnotesize
\begin{tcolorbox}[boxrule=0.5pt, top=0pt, bottom=0pt, left=0pt, right=0pt, title=Direct Answer Prompt]
Please read the multiple-choice question below carefully and select ONE of the listed options. Provide only the symbols and do not output anything else after that.\\
\\
Question: \{question\}\\
Options:\\
\{options\}\\
Answer: 
\end{tcolorbox}
\begin{tcolorbox}[boxrule=0.5pt, top=0pt, bottom=0pt, left=0pt, right=0pt, title=Chain-of-Thought Prompt]
Please read the multiple-choice question below carefully, explain step-by-step, and select ONE of the listed options. Provide only the explanations and symbols. Do not output anything else after that. \\
\\
Question: \{question\} \\
Options: \\
\{options\} \\
Answer: 
\end{tcolorbox}
\caption{Listing our prompts used as the direct answer (DA) and the zero-shot chain-of-thought (CoT).
The placeholders ``\{question\}'' and ``\{options\}'' are replaced with an actual question and its options, as in Figure~\ref{fig:example}.}
\label{fig:prompt}
\end{figure}

\section{Implementation Details}
\label{sec:implementation}
We incorporated self-consistency with the best performing MCQA instructions and the zero-shot CoT prompt by turning them into generalized pre- and post-processing steps for LLMs, as follows.

\paragraph{Pre-processing.}
We generate a LLM prompt using three types of instructions used in relevant studies:
\begin{inparaenum}[(a)]
\item the general instruction, e.g. ``\textit{Please read the multiple-choice question}'' \cite{wang-etal-2024-answer-c}, 
\item the negative instruction that prevents LLMs from producing unnecessary output, e.g. ``\textit{Provide only the symbols}'' \cite{robinson-etal-2023-chatgpt}, and
\item the zero-shot CoT prompt, e.g. ``\textit{explain step-by-step}'' \cite{kojima2022large}, which is used only when necessary.
\end{inparaenum}

\paragraph{Post-processing.}
After obtaining multiple samples from LLMs, we parse each sample to obtain candidate answers, and then use self-consistency \cite{wang2023selfconsistency} to determine the final answer.
This is done in three steps:
\begin{inparaenum}[(i)]
\item We parse the last line in the LLM output and extract the first valid answer.
For example, from \textit{``Option A. 3776m''} in Figure~\ref{fig:eyecatch}, we extract the option A.
Unlike \citet{wang2023selfconsistency} who used a string template ``The answer is X'', we assumed the alphabetical options [A-Z] only.
In the case of GSM8K, we extract the numeric answers instead.
\item We dismiss the samples from which we cannot extract a valid answer.
\item We perform majority vote over the extracted answers.
For instance, from the valid answers $\{\text{A}, \text{A}, \text{C}\}$ in Figure~\ref{fig:eyecatch}, we determine that the final answer is A.
\end{inparaenum}

\paragraph{Tie-Breaking.}
To ensure reproducibility, voting ties in our implementation (e.g. two votes for A versus two votes for B) are resolved deterministically by selecting the first option in alphabetical order (i.e. A).
The quantitative analysis in \S\ref{sec:quantitative} inherently accounts for such disagreement (e.g. $s=\frac{2}{2 + 2}=0.5$), appropriately reflecting the reduced confidence of tied votes.

\begin{table}[t]
\centering
\footnotesize
\begin{tabular}{@{}l@{}lccc@{}}
\toprule
& & \multicolumn{3}{c}{Accuracy (\%) $\uparrow$} \\
& & \multicolumn{3}{c}{MMLU dev} \\
\cmidrule(l){3-5}
Prompt & \multicolumn{1}{c}{Sampling} & \multicolumn{1}{c}{All} & Reasoning & Knowledge \\
\midrule
\multicolumn{5}{c}{\textsc{with 0-shot}} \\
CoT & \multicolumn{1}{c}{nucleus} & 80.35 & 84.44 & 78.46 \\
CoT & +\textsc{SC}~\scriptsize{($n$=5)} & \textbf{83.16} & \textbf{86.67} & \textbf{81.54} \\
CoT & +\textsc{SC}~\scriptsize{($n$=20)} & \textbf{82.81} & \textbf{88.89} & 80.00 \\
\midrule
\multicolumn{5}{c}{\textsc{with 4-shot}} \\
CoT & \multicolumn{1}{c}{nucleus} & 80.35 & 80.00 & 80.51 \\
CoT & +\textsc{SC}~\scriptsize{($n$=5)} & 80.35 & 78.89 & \textbf{81.03} \\
CoT & +\textsc{SC}~\scriptsize{($n$=20)} & \textbf{82.46} & \textbf{85.56} & \textbf{81.03} \\
\bottomrule
\end{tabular}
\caption{
Comparisons of our zero-shot method with the conventional few-shot method using GPT-4o-mini.
Bold text highlights improvements over the vanilla CoT baseline.
}\label{tab:vs-few}
\end{table}

\section{Zero-shot versus Few-shot}
\label{sec:vs-fewshot}
Unlike our zero-shot inference setting, \citet{chung2024scaling} performed experiments with a few-shot setting by using the MMLU development data that contains five examples for each subject as demonstration exemplars.
We conducted similar experiments on the MMLU development data.
To prevent data leakage for our assessment, we used four examples from each subject instead of five.

Table~\ref{tab:vs-few} summarizes our preliminary experiments comparing our zero-shot method with the conventional few-shot method on the MMLU development data.
These results demonstrate that our zero-shot method outperformed the previously used few-shot method, indicating its flexibility to parse arbitrary answers over a string template.
The only exception is a single instance ($n$=20), which can be attributed to the limited performance of GPT-4o-mini.

\begin{table}[t]
\centering
\footnotesize
\begin{tabular}{@{}c@{}lccc@{}}
\toprule
& & \multicolumn{3}{c}{Accuracy (\%) $\uparrow$} \\
& & \multicolumn{3}{c}{MMLU test} \\
\cmidrule(l){3-5}
Prompt & \multicolumn{1}{c}{Sampling} & \multicolumn{1}{c}{All} & Reasoning & Knowledge \\
\midrule
\multicolumn{5}{c}{\textsc{GPT-4o}} \\
DA & \multicolumn{1}{c}{nucleus} & 83.26 & 75.45 & 85.56 \\
CoT & \multicolumn{1}{c}{nucleus} & 87.86 & 90.38 & 87.12 \\
CoT & +\textsc{SC}~\scriptsize{($n$=5)} & \textbf{88.64}\rlap{$^*$} & \textbf{91.32}\rlap{$^*$} & \textbf{87.85}\rlap{$^*$} \\
CoT & +\textsc{SC}~\scriptsize{($n$=20)} & \textbf{88.93}\rlap{$^*$} & \textbf{91.94}\rlap{$^*$} & \textbf{88.04}\rlap{$^*$} \\
\midrule
\multicolumn{5}{c}{\textsc{GPT-4o-mini}} \\
DA & \multicolumn{1}{c}{nucleus} & 74.80 & 66.76 & 77.18 \\
CoT & \multicolumn{1}{c}{nucleus} & 81.43 & 84.41 & 80.56 \\
CoT & +\textsc{SC}~\scriptsize{($n$=5)} & \textbf{82.56}\rlap{$^*$} & \textbf{85.44}\rlap{$^*$} & \textbf{81.71}\rlap{$^*$} \\
CoT & +\textsc{SC}~\scriptsize{($n$=20)} & \textbf{82.61}\rlap{$^*$} & \textbf{85.57}\rlap{$^*$} & \textbf{81.74}\rlap{$^*$} \\
\midrule
\multicolumn{5}{c}{\textsc{Qwen2.5-32B-Instruct}} \\
DA & \multicolumn{1}{c}{nucleus} & 79.82 & 77.01 & 80.65 \\
CoT & \multicolumn{1}{c}{nucleus} & 80.06 & 82.66 & 79.29 \\
CoT & +\textsc{SC}~\scriptsize{($n$=5)} & \textbf{81.21}\rlap{$^*$} & \textbf{85.94}\rlap{$^*$} & \textbf{79.81} \\
CoT & +\textsc{SC}~\scriptsize{($n$=20)} & \textbf{82.90}\rlap{$^*$} & \textbf{87.13}\rlap{$^*$} & \textbf{81.65}\rlap{$^*$} \\
\bottomrule
\end{tabular}
\caption{
Performance on the MMLU test set and our proposed splits.
The asterisk ($^*$) denotes statistical significance ($p < 0.05$).
}\label{tab:internal-full}
\end{table}

\begin{table}[t]
\centering
\footnotesize
\begin{tabular}{@{}l@{}lcc@{}}
\toprule
& & \multicolumn{2}{c}{Accuracy (\%) $\uparrow$} \\
Prompt & \multicolumn{1}{c}{Sampling} & GSM8K & MedMCQA \\
\midrule
\multicolumn{3}{c}{\textsc{GPT-4o}} \\
DA & \multicolumn{1}{c}{nucleus} & 46.93 & 75.07 \\
CoT & \multicolumn{1}{c}{nucleus} & 84.23 & 76.76 \\
CoT & +\textsc{SC}~\scriptsize{($n$=5)} & \textbf{84.31} & \textbf{77.67}\rlap{$^*$} \\
CoT & +\textsc{SC}~\scriptsize{($n$=20)} & \textbf{84.46} & \textbf{77.41} \\
\midrule
\multicolumn{3}{c}{\textsc{GPT-4o-mini}} \\
DA & \multicolumn{1}{c}{nucleus} & 29.95 & 65.86 \\
CoT & \multicolumn{1}{c}{nucleus} & 86.13 & 67.51 \\
CoT & +\textsc{SC}~\scriptsize{($n$=5)} & 86.13 & \textbf{68.23} \\
CoT & +\textsc{SC}~\scriptsize{($n$=20)} & \textbf{86.88} & \textbf{68.66}\rlap{$^*$} \\
\bottomrule
\end{tabular}
\caption{
Performance on GSM8K (symbolic reasoning) and MedMCQA (knowledge recall).
These results on prototypical benchmarks further validate our MMLU split.
}\label{tab:external-full}
\end{table}

\section{Full Results}
\label{sec:full}

This section presents the full results of our experiments.
Consistent with our main findings, the results show that self-consistency generally improves performance over vanilla CoT.
Table~\ref{tab:internal-full} details the performance of GPT-4o, GPT-4o-mini, and Qwen2.5-32B-Instruct on the MMLU test
set and our proposed splits.
Table~\ref{tab:external-full} shows the corresponding results for GPT-4o and GPT-4o-mini on the GSM8K and MedMCQA benchmarks.

\section{Data-driven Data Splitting}
\label{sec:data-driven}
As an alternative to the heuristic-based split used in our main analysis, we also explored a pure data-driven data split.
Following an analysis by \citet{sprague2025to}, we ranked all MMLU subjects by the performance gain from CoT over a direct answer baseline, using results from GPT-4o.
This model-dependent ranking, shown in Table~\ref{tab:data-driven}, contrasts with our primary split, which applies a heuristic in a model-independent manner.

To quantify the agreement between the two splits, we calculated the Area Under the Receiver Operating Characteristic Curve \citep[AUC;][]{fawcett2006aucroc} and achieved a score of 0.96, indicating a strong overlap.
The ranking confirms that math subjects such as High School Mathematics enjoyed the most benefit from CoT, while humanity subjects such as High School History resulted in modest improvements.
The notable discrepancies with our heuristic-based split include subjects such as Medicine, Management, Nutrition, and Biology.

\begin{table}[t]
\centering
\footnotesize
\begin{tabular}{@{}c@{}c@{}}
\toprule
Subject & $\Delta$ CoT $-$ DA \\
\midrule
High School Mathematics & 40.37 \\
Abstract Algebra & 24.00 \\
College Physics & 23.53 \\
College Mathematics & 23.00 \\
Elementary Mathematics & 21.96 \\
Formal Logic & 19.84 \\
Professional Accounting & 19.51 \\
College Chemistry & 16.00 \\
Moral Scenarios & 14.30 \\
College Computer Science & 14.00 \\
High School Chemistry & 13.30 \\
Econometrics & 13.16 \\
High School Statistics & 12.50 \\
High School Physics & 10.59 \\
Machine Learning & 8.04 \\
High School Computer Science & 8.00 \\
College Medicine & 5.78 \\
Management & 3.88 \\
High School Microeconomics & 3.36 \\
High School Macroeconomics & 3.34 \\
Nutrition & 3.27 \\
Conceptual Physics & 2.98 \\
Electrical Engineering & 2.76 \\
Astronomy & 2.63 \\
College Biology & 2.09 \\
Business Ethics & 2.00 \\
Sociology & 1.49 \\
Anatomy & 1.48 \\
\midrule
High School US History & 1.47\rlap{$^\sim$} \\
\midrule
Miscellaneous & 1.40 \\
High School Government and Politics & 1.03 \\
Medical Genetics & 1.00 \\
Professional Law & 0.91 \\
Professional Psychology & 0.66 \\
Logical Fallacies & 0.62 \\
High School Psychology & 0.37 \\
High School Biology & 0.32 \\
Computer Security & 0.00 \\
Human Aging & 0.00 \\
Professional Medicine & 0.00 \\
Prehistory & -0.30 \\
Clinical Knowledge & -0.38 \\
Marketing & -0.42 \\
International Law & -0.82 \\
Jurisprudence & -0.92 \\
US Foreign Policy & -1.00 \\
World Religions & -1.17 \\
High School European History & -1.22 \\
Security Studies & -1.63 \\
Virology & -1.81 \\
Public Relations & -1.81 \\
Moral Disputes & -2.32 \\
Human Sexuality & -3.05 \\
High School Geography & -3.53 \\
Philosophy & -3.86 \\
High School World History & -4.22 \\
Global Facts & -5.00 \\
\bottomrule
\end{tabular}
\caption{
Data-driven splitting obtained for the MMLU test using GPT-4o.
The tilde ($\sim$) denotes the median value, which corresponds to High School US History.
}\label{tab:data-driven}
\end{table}

\end{document}